\documentclass[runningheads]{llncs}
\usepackage[T1]{fontenc}

\usepackage{graphicx}
\usepackage{cite}
\usepackage{amsmath,amssymb,amsfonts}
\usepackage{algorithmic}
\usepackage{graphicx}
\usepackage{float}

\usepackage{algorithmic}
\usepackage[ruled,vlined]{algorithm2e}
\usepackage{cuted, nccmath}
\usepackage{caption}
\usepackage{subcaption}
\usepackage{textcomp}
\usepackage{xcolor}
\usepackage{subfiles}
\usepackage{comment}

\begin{document}

\title{Towards multi-modal anatomical landmark detection for ultrasound-guided brain tumor resection with contrastive learning}

\author{Soorena Salari 
\inst{1} \and
Amirhossein Rasoulian 
\inst{1}\and
Hassan Rivaz
\inst{2}\and
Yiming Xiao
\inst{1}}
\authorrunning{S. Salari et al.}
%
\institute{Department of Computer Science and Software Engineering, Concordia University, Montreal, Canada\\
\email{\{soorena.salari,ah.rasoulian,yiming.xiao\}@concordia.ca}
\and
Department of Electrical and Computer Engineering, Concordia University, Montreal, Canada\\
\email{hassan.rivaz@concordia.ca}}

\maketitle              

\begin{abstract}
Homologous anatomical landmarks between medical scans are instrumental in quantitative assessment of image registration quality in various clinical applications, such as MRI-ultrasound registration for tissue shift correction in ultrasound-guided brain tumor resection. While manually identified landmark pairs between MRI and ultrasound (US) have greatly facilitated the validation of different registration algorithms for the task, the procedure requires significant expertise, labor, and time, and can be prone to inter- and intra-rater inconsistency. So far, many traditional and machine learning  approaches have been presented for anatomical landmark detection, but they primarily focus on mono-modal applications. Unfortunately, despite the clinical needs, inter-modal/contrast landmark detection has very rarely been attempted. Therefore, we propose a novel contrastive learning framework to detect corresponding landmarks between MRI and intra-operative US scans in neurosurgery. Specifically, two convolutional neural networks were trained jointly to encode image features in MRI and US scans to help match the US image patch that contain the corresponding landmarks in the MRI. We developed and validated the technique using the public RESECT database. With a mean landmark detection accuracy of 5.88±4.79 mm against 18.78±4.77 mm with SIFT features, the proposed method offers promising results for MRI-US landmark detection in neurosurgical applications for the first time. 

\end{abstract}

\section{Introduction}
Gliomas are the most common central nervous system (CNS) tumors in adults, accounting for 80$\%$ of primary malignant brain tumors \cite{holland2001progenitor}. Early surgical treatment to remove the maximum amount of cancerous tissues while preserving the eloquent brain regions can improve the patient's survival rate and functional outcomes of the procedure \cite{dolecek2012cbtrus}. Although the latest multi-modal medical imaging (e.g, PET, diffusion/functional MRI) allows more precise pre-surigcal planning, during surgery, brain tissues can deform under multiple factors, such as gravity, intracranial pressure change, and drug administration. The phenomenon is referred to as brain shift, and often invalidates the pre-surgical plan by displacing surgical targets and other vital anatomies. With high flexibility, portability, and cost-effectiveness, intra-operative ultrasound (US) is a popular choice to track and monitor brain shift. In conjunction with effective MRI-US registration algorithms, the tool can help update the pre-surgical plan during surgery to ensure the accuracy and safety of the intervention.

As the true underlying deformation from brain shift is impossible to obtain and the differences of image features between MRI and US are large, quantitative validation of automatic MRI-US registration algorithms often rely on homologous anatomical landmarks that are manually labeled between corresponding MRI and intra-operative US scans  \cite{xiao2019evaluation}. However, manual landmark identification requires strong expertise in anatomy and is costly in labor and time. Moreover, inter- and intra-rater variability still exists. These factors make quality assessment of brain shift correction for US-guided brain tumor resection challenging. In addition, due to the time constraints, similar evaluation of inter-modal registration quality during surgery is nearly impossible, but still highly desirable. To address these needs, deep learning (DL) holds the promise to perform efficient and automatic inter-modal anatomical landmark detection. 

Previously, many groups have proposed algorithms to label landmarks in anatomical scans \cite{yao2021label,ghesu2016artificial,zhu2021you,tripathi2023unsupervised,toews2013efficient,salari2023uncertainty}. However, almost all earlier techniques were designed for mono-modal applications, and inter-modal landmark detection, such as for US-guided brain tumor resection, has rarely been attempted. In addition, unlike other applications, where the full anatomy is visible in the scan and all landmarks have consistent spatial arrangements across subjects, intra-operative US of brain tumor resection only contains local regions of the pathology with non-canonical orientations. This results in anatomical landmarks with different spatial distributions across cases. To address these unique challenges, we proposed a new contrastive learning (CL) framework to detect matching landmarks in intra-operative US with those from MRI as references. Specifically, the technique leverages two convolutional neural networks (CNNs) to learn features between MRI and US that distinguish the inter-modal image patches which are centered at the matching landmarks from those that are not. Our approach has two major novel contributions to the field. \underline{First}, we proposed a multi-modal landmark detection algorithm for US-guided brain tumor resection for the first time. \underline{Second}, CL is employed for the first time in inter-modal anatomical landmark detection. We developed and validated the proposed technique with the public RESECT database \cite{xiao2017re} and compared its landmark detection accuracy against the popular scale-invariant feature transformation (SIFT) algorithm in 3D \cite{rister2017volumetric}.

\section{Related Work}

Contrastive learning  has recently shown great results in a wide range of medical image analysis tasks \cite{you2022intra,cheng2022contrastive,bhattacharya2022supervised,emre2022tinc,liu2022joint,pan2022vision,hang2022reliability}. In short, it seeks to boost the similarity of feature representations between counterpart samples and decrease those between mismatched pairs. Often, these similarities are calculated based on deep feature representations obtained from DL models in the feature embedding space. This self-supervised learning set-up allows robust feature learning and embedding without explicit guidance from fine-grained image annotations, and the encoded features can be adopted in various downstream tasks, such as segmentation. A few recent works \cite{quan2022images,quan2022information,yao2021one} explored the potential of CL in anatomical landmark annotation in head X-ray images for 2D skull landmarks. Quan et al. \cite{quan2022images,quan2022information} attempted to leverage CL for more efficient and robust learning. Yao et al. \cite{yao2021one} used multiscale pixel-wise contrastive proxy tasks for skull landmark detection in X-ray images. With a consistent protocol for landmark identification, they trained the network to learn signature features within local patches centered at the landmarks. These prior works with CL focus on single-modal 2D landmark identification with systematic landmark localization protocols and sharp image contrast (i.e., skull in X-ray). In contrast, our described application is more challenging due to the 3D nature, difficulty in inter-modal feature learning, weaker anatomical contrast (i.e., MRI vs US), and variable landmark locations. In CL, many works have employed the InfoNCE loss function \cite{gutmann2010noise,oord2018representation} in attaining good outcomes. Inspired by Yao \textit{et al.} \cite{yao2021one}, we aimed to use InfoNCE as our loss function with a patch-based approach. To date, CL has not been explored in multi-modal landmark detection, a unique problem in clinical applications. In this paper, to bridge this knowledge gap, we proposed a novel CL-based framework for MRI-US anatomical landmark detection.

\section{Methods and Materials}
\subsection{Data and landmark annotation}
\begin{figure}[htbp]
\centering
\centerline{\includegraphics[scale=0.28]{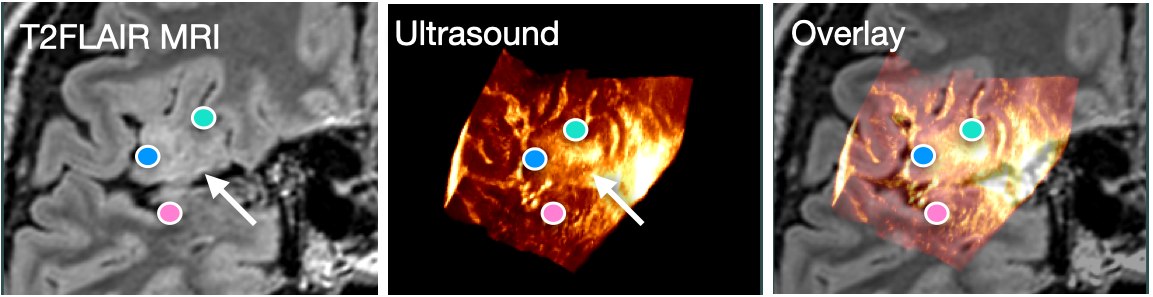}}
\caption{Sample corresponding landmarks on co-registered T2FLAIR MRI and US. The arrows point to the brain tumor region.}
\label{SampleLandmarks}
\end{figure}

We employed the publicly available EASY-RESECT (REtroSpective Evaluation of Cerebral Tumors) dataset \cite{xiao2017re} (https://archive.sigma2.no/pages/public/dataset Detail.jsf?id=10.11582/2020.00025) to train and evaluate our proposed method. This dataset is a deep-learning-ready version of the original RESECT database, and was released as part of the 2020 Learn2Reg Challenge \cite{hering2022learn2reg}. Specifically, EASY-RESECT contains MRI and intra-operative US scans (before resection) of 22 subjects who have undergone low-grade glioma resection surgeries. All images were resampled to a unified dimension of 256 $\times$ 256 $\times$ 288 voxels, with an isotropic resolution of $\sim$0.5mm. Between MRI and the corresponding US images, matching anatomical landmarks were manually labeled by experts and 15$\sim$16 landmarks were available per case. A sample illustration of corresponding inter-modal scans and landmarks is shown in Fig. \ref{SampleLandmarks}. For the target application, we employed the T2FLAIR MRI to pair with intra-operative US since low-grade gliomas are usually more discernible in T2FLAIR than in T1-weighted MRI\cite{xiao2017re}.

\subsection{Contrastive learning framework}
\begin{figure*}[h]
\includegraphics[scale=0.26]{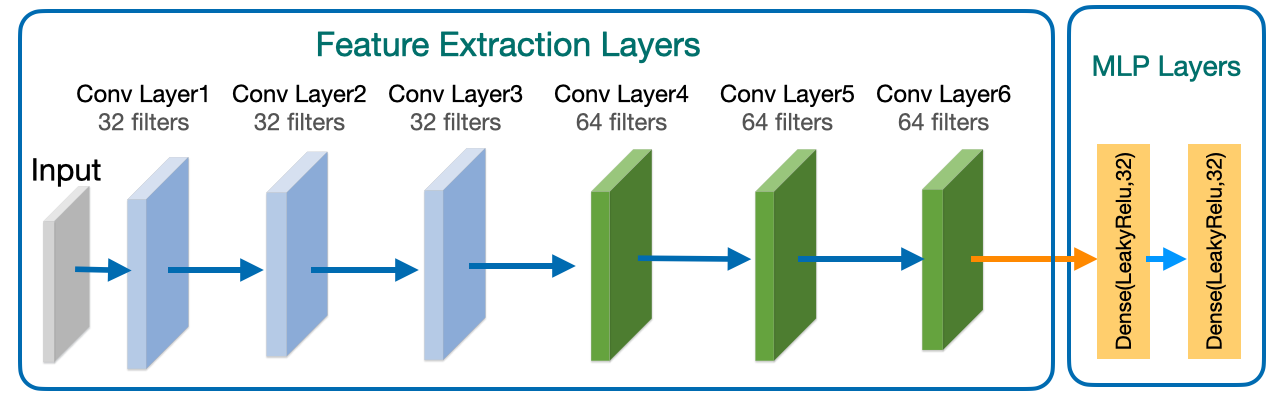}
\caption{The proposed CNN for feature encoding from MRI and US scans.}
\label{CNN}
\end{figure*}

We used two  CNNs with identical architectures in parallel to extract robust image features from MRI and US scans. Specifically, these CNNs are designed to acquire relevant features from MRI and US patches, and maximize the similarity between features of corresponding patches while minimizing those between mismatched patches. Each CNN network contains six successive blocks, and each block consists of one convolution layer and one group norm, with Leaky ReLU as the activation function. Also, the convolution layer of the first and last three blocks of the network has 64 and 32 convolutional filters, respectively, and a kernel size of 3 is used across all blocks. After the convolution layers, the proposed network has two multi-layer perceptron (MLP) layers with 64 and 32 neurons and Leaky ReLU as the activation function. These MLP layers compress the extracted features from convolutional layers and produce the final feature vectors. The resulting CNN network is depicted in Fig. \ref{CNN}.

\subsection{Landmark matching with a 2.5D approach}
\label{2.5D}
\begin{figure*}[htpb]
\centering
\includegraphics[scale=0.35
]{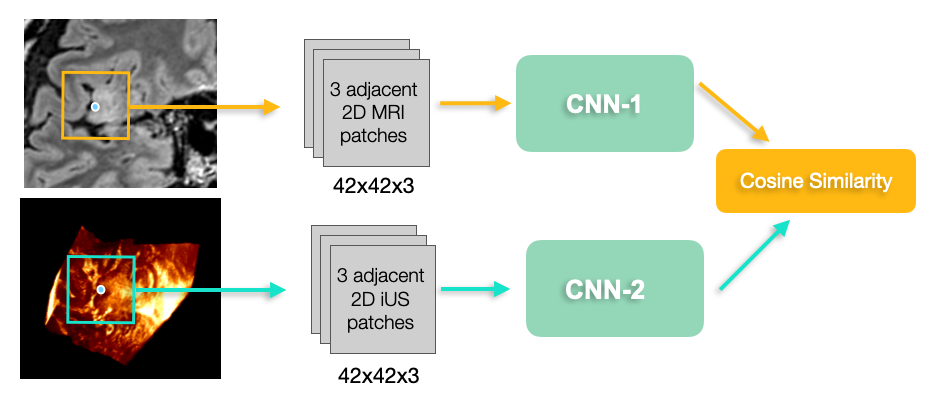}
\caption{An overview of the framework for image feature learning.}
\label{Framework}
\end{figure*}

Working with 3D images is computationally expensive and can make the model training unstable and prone to overfitting, especially when the size of the database is limited. Therefore, instead of a full 3D processing, we decided to implement a 2.5D approach \cite{pirhadi2022robust} to leverage the efficiency of 2D CNN in the CL framework for the task. In this case, we extracted a series of three adjacent 2D image patches in one canonical direction (x-, y-, or z-direction), with the middle slice centred at the true or candidate landmarks in a 3D scan to provide slight spatial context for the middle slice of interest. To construct the full 2.5D formulation, we performed the same image patch series extraction in all x-, y-, and z-directions for a landmark, and this 2.5D patch forms the basis to compute the similarity between the queried US and reference MRI patches.  Note that the setup of CL requires three types of samples, anchor, positive sample pairs, and negative sample pairs. Specifically, the anchor is defined as the 2.5D MRI patch centred at a predefined landmark, a positive pair is represented by an anchor and a 2.5D US patch at the corresponding landmark, and finally, a negative pair means an anchor and a mismatched 2.5D US patch. Note that during network training, instead of 2.5D patches, we compared the 2D image patch series in one canonical direction between MRI and US, and 2D patch series in all three directions were used. During the inference stage, the similarity between MRI and US 2.5D patches was obtained by summing the similarities of corresponding 2D image patch series in each direction, and a match was determined with the highest similarity from all queried US patches.
With the assumption that the brain shift moves the anatomy within a limited range, during the inference time, we searched within a range of [-5,5] mm in each direction in the US around the reference MRI landmark location to find the best match. Note that this search range is an adjustable parameter by the user (e.g., surgeons/clinicians), and when no match is found in the search range, an extended search range can be used. The general overview of the utilized framework for 2D image patch extraction is shown in Fig. \ref{Framework}.

\subsection{Landmark matching with 3D SIFT}

The SIFT algorithm \cite{rister2017volumetric} is a well-known tool for keypoint detection and image registration. It has been widely used in multi-modal medical registration, such as landmark matching for brain shift correction in image-guided neurosurgery \cite{luo2018feature,toews2013efficient}. To further validate the proposed CL-based method for multi-modal anatomical landmark detection in US scans, we replicated the procedure using the 3D SIFT algorithm as follows. First, we calculated the SIFT features at the reference landmark's location in MRI. Then, we acquired a set of candidate SIFT points in the corresponding US scan. Finally, we identified the matching US landmark by selecting the top ranking candidate based on SIFT feature similarity measured with cosine similarity. Note that, for SIFT-based landmark matching, we have attempted to impose a similar spatial constraint like in the CL-based approach. However, as the SIFT algorithm pre-selects keypoint candidates based on their feature strengths, with this in consideration, we saw no major benefits by imposing the spatial constraint.

\section{Experimental setup}
\subsection{Data preprocessing}
For CL training, both positive and negative sample pairs need to be created. All 2D patch series were extracted according to Section \ref{2.5D} with a size of 42 $\times$ 42 $\times$ 3 voxels. These sample pairs were used to train two CNNs to extract relevant image features across MRI and US leveraging the InfoNCE loss.

\subsection{Loss function}
We used the InfoNCE loss \cite{oord2018representation} for our CL framework. The loss function has been widely used and demonstrated great performance in many vision tasks. Like other contrastive loss functions, InfoNCE requires a similarity function, and we chose commonly used cosine similarity. The formulas for InfoNCE ($L_{InfoNCE}$) and cosine similarity ($CosSim$) are as follows:

\begin{equation}
\begin{aligned}
\mathcal{L}_{\text {InfoNCE }} &=-\mathbb{E}\left[\log \frac{\exp (\alpha)}{\exp (\alpha)+\sum \exp \left(\alpha^{\prime}\right)}\right] ; \\
\alpha &= s [F_\theta \circ X_i^A, G_\beta \circ X_i^P] ; \\
\alpha^{\prime} &= s [F_\theta \circ X_i^A, G_\beta \circ X_j^N],
\end{aligned}
\end{equation}

\begin{equation}
s\left[v, w\right]=\operatorname{Cos} \operatorname{Sim}\left(v, w\right)=\frac{\left\langle v \cdot w\right\rangle}{\left\|v\right\| \cdot\left\|w\right\|}
\end{equation}

\noindent where $F_\theta$ and $G_\beta$ are the CNN feature extractors for MR and US patches. $X_i^A$ and $X_i^P$ are the cropped image patches around the corresponding landmarks in MR and US scans, respectively, and $X_i^N$ is a mismatched patch in the US image to that cropped around the MRI reference landmark. Here, $F_\theta \circ X_i^A$, $G_\beta \circ X_i^P$, and $G_\beta \circ X_j^N$ give the extracted feature vectors for MR and US patches.

\subsection{Implementation details and evaluation}
To train our DL model, we made subject-wise division of the entire dataset into 70\%:15\%:15\% as the training, validation, and testing sets, respectively. Also, to improve the robustness of the network, we used data augmentation for the training data by random rotation, random horizontal flip, and random vertical flip. Furthermore, an AdamW optimizer with a learning rate of 0.00001 was used, and we trained our model for 50 epochs with a batch size of 256.

In order to evaluate the performance of our technique, we used the provided ground truth landmarks from the database and calculated the Euclidean distance between the ground truths and predictions. The utilized metric is as follows:

\begin{equation}
\text{Mean landmark identification error} =\frac{1}{N} \sum_{i=1}^N\left\|x_i-x_i^{\prime}\right\|
\end{equation}

\noindent where $x_i$ and $x_i^{\prime}$, and $N$ are the ground truth landmark location, model prediction, and the total number of landmarks per subject, respectively. 

\section{Results}

Table \ref{Comparison_backbones} lists the mean and standard deviation of landmark identification errors (in mm) between the predicted position and the ground truth in intra-operative US for each patient of the RESECT dataset. In the table, we also provide the severity of brain shift for each patient. Here, tissue deformation measured as mean target registration errors (mTREs) with the ground truth anatomical landmarks is classified as small (mTRE below 3 mm), median (3-6 mm), or large (above 6mm). The results show that our CL-based landmark selection technique can locate the corresponding US landmarks with a mean landmark identification error of 5.88$\pm$4.79 mm across all cases while the SIFT algorithm has an error 18.78$\pm$4.77 mm. With a two-sided paired-samples t-test, our method outperformed the SIFT approach with statistical significance ($p<$1e-4). When reviewing the mean landmark identification error using our proposed technique, we also found that the magnitude is associated with the level of brain shift. However, no such trend is observed when using SIFT features for landmark identification. When inspecting landmark identification errors across all subjects between the CL and SIFT techniques, we also noticed that our CL framework has significantly lower standard deviations ($p<$1e-4), implying that our technique has a better performance consistency.

\begin{table}[]

\centering

\caption{Landmark identification errors (mean$\pm$std) per case in mm. Our proposed CL-based algorithm achieved a mean landmark identification error of \textbf{5.88}$\pm$\textbf{4.79} mm across all cases while the SIFT algorithm obtained an error of \textbf{18.78}$\pm$\textbf{4.77} mm. The level of brain shift is listed beside the patient ID.}
\begin{tabular}{c c c c c c p{4cm}p{4.5cm}p{4.5cm}p{4.5cm}p{4.5cm}p{4.5cm} p{4cm}p{4.5cm}p{4.5cm}p{4.5cm}p{4.5cm}p{4.5cm}}
\hline
 Patient ID  & Proposed CL & SIFT Algorithm  & Patient ID  & Proposed CL & SIFT Algorithm \\

\hline
\hline
1 (Small) &  1.80$\pm$0.78 & 12.16$\pm$4.75 & 15 (Medium) &  3.07$\pm$1.39 & 19.33$\pm$4.49\\
\hline

2 (Medium) &  6.16$\pm$1.40 & 17.84$\pm$8.50 & 16 (Medium) &  6.42$\pm$0.92 & 12.48$\pm$4.84 \\
\hline

3 (Large) &  7.79$\pm$0.55 & 13.37$\pm$6.08 & 17 (Large) &  8.13$\pm$0.72 & 18.57$\pm$6.27\\
\hline

4 (Small) &  3.59$\pm$0.73 & 13.97$\pm$5.50 & 18 (Medium) &  4.19$\pm$0.87 & 13.71$\pm$5.815\\
\hline

5 (Large) &  10.65$\pm$1.13 & 20.71$\pm$6.12 & 19 (Medium) &  3.97$\pm$0.93 & 27.70$\pm$16.49\\
\hline

6 (Medium) &  2.20$\pm$0.93 & 28.11$\pm$11.90 & 21 (Medium) &  6.01$\pm$0.77 & 24.40$\pm$13.73\\
\hline

7 (Small) &  1.96$\pm$0.96 & 24.07$\pm$8.02 & 23 (Large)&  6.97$\pm$1.03 & 22.50$\pm$6.72\\
\hline

8 (Small) &  2.56$\pm$0.79 & 19.30$\pm$6.09 & 24 (Small)&  1.33$\pm$0.49 & 14.91$\pm$6.09\\
\hline

12 (Large) &  23.77$\pm$0.96 & 22.01$\pm$6.64 & 25 (Large) &  9.94$\pm$2.43 & 15.37$\pm$5.42\\
\hline

13 (Medium) &  6.18$\pm$1.43 & 13.86$\pm$6.99 & 26 (Small) &  2.95$\pm$0.88 & 17.93$\pm$10.15\\
\hline

14 (Medium) &  3.39$\pm$0.69 & 21.67$\pm$6.46 & 27 (Medium) &  6.42$\pm$0.76 & 19.20$\pm$8.65\\
\hline

\hline

\end{tabular}
\label{Comparison_backbones}
\end{table}

\section{Discussion}
Inter-modal anatomical landmark localization is still a difficult task, especially for the described application, where landmarks have no consistent spatial arrangement across different cases and image features in US are rough. We tackled the challenge with the CL framework for the first time. As the first step towards more accurate inter-modal landmark localization, there are still aspects to be improved. First, while the 2.5D approach is memory efficient and quick, 3D approaches may better capture the full corresponding image features. This is partially reflected by the observation that the quality of landmark localization is associated with the level of tissue shift. However, due to limited clinical data, 3D approaches caused overfitting in our network training.  Second, in the current setup, we employed landmarks in pre-operative MRIs as references since its contrast is easier to understand and it allows sufficient time for clinicians to annotate the landmarks before surgery. Future exploration will also seek techniques to automatically tag MRI reference landmarks. Finally, we only employed US scans before resection since tissue removal can further complicate feature matching between MRI and US, and requires more elaborate strategies, such as those involving segmentation of resected regions \cite{canalini2019segmentation}. We will explore suitable solutions to extend the application scenarios of our proposed framework as part of the future investigation. As a baseline comparison, we employed the SIFT algorithm, which has demonstrated excellent performance in a large variety of computer vision problems for keypoint matching. However, in the described inter-modal landmark identification for US-guided brain tumor resection, the SIFT algorithm didn't offer satisfactory results. This could be due to the coarse image features and textures of intra-operative US and the differences in the physical resolution between MRI and US. One major critique for using the SIFT algorithm is that it intends to find geometrically interesting keypoints, which may not have good anatomical significance. In the RESECT dataset, eligible anatomical landmarks were defined as deep grooves and corners of sulci, convex points of gyri, and vanishing points of sulci. The relevant local features may be hard to capture with the SIFT algorithm. In this sense, DL-based approaches may be a better choice for the task. With the CL framework, our method learns the common features between two different modalities via the training process. Besides better landmark identification accuracy, the tighter standard deviations also imply that our DL approach serves a better role in grasping the local image features within the image patches. 

\section{Conclusions}

In this project, we proposed a CL framework for MRI-US landmark detection for neurosurgery for the first time by leveraging real clinical data, and achieved state-of-the-art results. The algorithm represents the first step towards efficient and accurate inter-modal landmark identification that has the potential to allow intra-operative assessment of registration quality. Future extension of the method in other inter-modal applications can further confirm its robustness and accuracy.
\\ \\
\noindent \textbf{Acknowledgment}. We acknowledge the support of the Natural Sciences and Engineering Research Council of Canada (NSERC) and Fonds de Recherche du Québec Nature et technologies (FRQNT).

\bibliographystyle{IEEEtran}
\bibliography{ref.bib}

\end{document}